\documentclass{article}
\usepackage{arxiv}

\usepackage[utf8]{inputenc} % allow utf-8 input
\usepackage[T1]{fontenc}    % use 8-bit T1 fonts
\usepackage{hyperref}       % hyperlinks
\usepackage{url}            % simple URL typesetting
\usepackage{booktabs}       % professional-quality tables
\usepackage{amsfonts}       % blackboard math symbols
\usepackage{nicefrac}       % compact symbols for 1/2, etc.
\usepackage{microtype}      % microtypography
\usepackage{lipsum}		% Can be removed after putting your text content
\usepackage{graphicx}
\usepackage[numbers]{natbib}
\usepackage{amsmath, amssymb,amsthm}
\usepackage{doi}
\usepackage{algorithmic}
\usepackage{subfig}
\usepackage{dblfloatfix}
\usepackage{algorithm,algorithmic}
\usepackage{textcomp}
\usepackage{makecell}
\usepackage{array} 
\usepackage{titlesec}

\hypersetup{
hidelinks,
pdftitle={A template for the arxiv style},
pdfsubject={q-bio.NC, q-bio.QM},
pdfauthor={Pranshu~Jena, Prajnyajit~Mohanty, Umesh~C.~Pati},
pdfkeywords={First keyword, Second keyword, More},
}

\title{FedTumor-IoMT: Federated Learning-Aided Brain Tumor Diagnosis on Resource-Constrained Edge IoMT Devices}

\author{
\begin{minipage}[t]{0.32\textwidth}
\centering
\href{https://orcid.org/0009-0007-1518-8973}
{\includegraphics[height=0.9em]{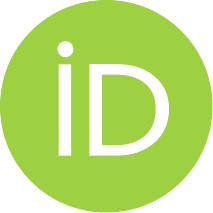}\hspace{1mm}\textbf{Pranshu Jena}}\\
Department of ECE\\
National Institute of Technology Rourkela\\
India\\
\texttt{\scriptsize pranshujena16@gmail.com}
\end{minipage}
\hfill
\begin{minipage}[t]{0.32\textwidth}
\centering
\href{https://orcid.org/0000-0002-5653-3602}
{\includegraphics[height=0.9em]{orcid.pdf}\hspace{1mm}\textbf{Prajnayjit Mohanty}}\\
School of Electrical and Computer Sciences\\
Indian Institute of Technology Bhubaneswar\\
India\\
\texttt{\scriptsize prajnyajit@iitbbs.ac.in}
\end{minipage}
\hfill
\begin{minipage}[t]{0.32\textwidth}
\centering
\href{https://orcid.org/0000-0001-9805-2543}
{\includegraphics[height=0.9em]{orcid.pdf}\hspace{1mm}\textbf{Umesh C. Pati}}\\
Department of ECE\\
National Institute of Technology Rourkela\\
India\\
\texttt{\scriptsize ucpati@nitrkl.ac.in}
\end{minipage}
}

\renewcommand{\shorttitle}{FedTumor-IoMT for Brain Tumor Diagnosis}

\hypersetup{
pdftitle={A template for the arxiv style},
pdfsubject={q-bio.NC, q-bio.QM},
pdfauthor={Pranshu~Jena, Prajnyajit~Mohanty, Umesh~C.~PAti},
pdfkeywords={First keyword, Second keyword, More},
}

\begin{document}
\maketitle

\begin{abstract}
	Joint brain tumor segmentation and classification models are computationally intensive, limiting their training on resource-constrained edge platforms and restricting privacy-preserving Federated Learning FL in the Internet of Medical Things IoMT. Existing approaches either perform both tasks on GPU-equipped workstations or support only classification on edge devices. To address this gap, this work proposes a lightweight multitask FL framework for joint segmentation and classification through local edge training. A shared pretrained LiteMedSAM encoder with TinyViT extracts features for a coordinate-attention-based segmentation decoder and a segmentation-guided classification head, eliminating a separate feature extractor. Selective encoder freezing retains 95\% trainable parameters while reducing on-device training costs. The proposed framework is experimentally validated on a laboratory testbed of five resource-constrained edge clients using 1,200 T1-weighted MRI samples. It achieves 94.20\% classification accuracy, 88.53\% DSC, 83.68\% IoU, and 99.87\% ROC-AUC. Its classification accuracy remains within 0.6\% of the GPU-trained baseline of 95.21\% across five FL rounds. The attention and feature refinement modules improve DSC and IoU by 3.82\% and 4.60\%, respectively, with 0.74 GFLOPs of additional computation. On edge hardware, the framework achieves 46.54 s inference latency, 52.8\% CPU utilization, and a peak temperature of 65$^\circ$C without thermal throttling. Its inference latency is observed to be 6.4--10.5 times lower than that of existing frameworks. These results demonstrate the feasibility of privacy-preserving joint brain tumor segmentation and classification on resource-constrained edge platforms.
\end{abstract}

% keywords can be removed

\section{Introduction}
% \lipsum[2]
% \lipsum[3]
\label{sec:introduction}
Brain tumors are life-threatening diseases that essentially demand early and accurate diagnosis for effective neuro-oncological treatment and improved clinical outcomes, particularly for gliomas, meningiomas, and pituitary tumors \cite{jbhi2,jbhi1}. The diagnosis involves tumor segmentation to delineate the region and classification to identify its category. These tasks are mutually dependent, as precise spatial delineation of the lesion directly informs its classification. While centralized training is effective, aggregating multi-source MRI scans introduces severe privacy and regulatory challenges \cite{jbhi3,jbhi4}. Additionally, transferring high-dimensional data to centralized servers incurs heavy bandwidth requirements, high latency, and security vulnerabilities during wireless communication \cite{resourceconstrained}. The Internet of Medical Things (IoMT) connects medical imaging systems and low-cost computing devices across healthcare facilities, enabling data processing closer to its source. It aligns with Industry 5.0 priorities of human-centricity, sustainability, and resilience. In this context, Federated Learning (FL) offers a practical approach to edge-assisted distributed learning, wherein hospitals train models locally and exchange model updates rather than raw MRI scans. This is particularly relevant because individual hospitals often possess limited, site-specific datasets, while data protection laws, and institutional policies restrict centralized data sharing. FL also allows models to accommodate variations in patient populations and imaging systems across hospitals. \\

However, FL places a training burden on edge devices, where limited memory, CPU capacity, and energy make deploying large Convolutional Neural Networks (CNNs) difficult. Existing joint segmentation--classification methods further exacerbate these challenges. Most state-of-the-art works use separate networks for the two tasks, leading to redundant feature extraction as well as increased computational and parameter costs \cite{KUMARSAHOO2023616,akter}. Although encoder-sharing multi-task models improve integration, recent hybrid designs have grown to 37~M--109~M parameters and run only on GPUs \cite{renugadevi2025hybrid, SALAMA2026433}. Conversely, the few edge-deployable models achieve small footprints by omitting segmentation entirely and support only inference, while requiring GPU-based training \cite{neena2026haar}. Moreover, state-of-the-art work on medical diagnosis has not sufficiently reported practical post-federated-learning deployment performance metrics for joint segmentation and classification. Crucially, essential deployment metrics, such as inference latency, memory footprint, and thermal behavior, remain largely undocumented for resource-constrained IoMT devices. This leaves a critical gap between model-level performance and real-world deployability, as the feasibility of local training and execution on low-cost edge hardware has not been sufficiently demonstrated. Addressing this gap is essential to translating the Industry 5.0 principles of privacy, sustainability, and resilience into practical medical Artificial Intelligence (AI) systems.

The major contributions are mentioned as follows:
%#####################################
%####To be written in end#############
%#####################################
%#####################################
\begin{itemize}
    \item An FL-aided IoMT framework is proposed that trains joint brain tumor segmentation and classification locally on resource-constrained edge clients, unlike state-of-the-art works, which rely entirely on GPU training to perform this joint task. A single shared LiteMedSAM encoder extracts features once and reuses them for both tasks, without an additional feature extractor, thereby substantially reducing computation and communication payloads and making the framework suitable for resource-constrained edge clients.
    
    \item The proposed framework introduces a coordinate-attention segmentation decoder with progressive channel reduction and a $1\times1$ convolution-based feature-refinement module, boosting the DSC by 3.82\% and IoU by 4.60\% at an additional 0.74 GFLOPs of computation.

    \item It applies selective encoder freezing to reduce on-device training computation while maintaining model performance across federated communication rounds, ensuring data privacy by exchanging only model weights rather than raw local data.
  
    \item The proposed framework has been experimentally evaluated in laboratory environment. It is trained on resource-constrained embedded computing client platform without thermal throttling, unlike GPU-dependent frameworks that fail or degrade on edge hardware. Compared to the state-of-the-art works on brain tumor detection, the proposed IoMT framework reduces CPU utilization to 52.8\% versus 68.5--99\%, and lowers per-image latency by $6.4$ to $10.5$ times, while achieving the highest accuracy among frameworks that run on the edge device.  
\end{itemize}
% Organization of the Paper to be written%

The remainder of this paper is organized as follows: Section \ref{rw} provides a detailed analysis of the existing literature on deep learning-based brain tumor segmentation and classification. The research gaps that motivate the present work are also identified there. Section \ref{overview} describes the proposed framework that covers the overall system architecture. Section \ref{result} presents the experimental setup as well as reports and discusses the results. Finally, Section \ref{conclusion} concludes the paper and outlines directions for future research.

\section{Related Prior Work} \label{rw}

Brain tumor analysis from MRI involves segmentation, which delineates the tumor region, and classification, which assigns the tumor type. In task-specific two-stage pipelines, both tasks are performed in sequence by separately trained networks as proposed in \cite{KUMARSAHOO2023616},\cite{jabbar}, \cite{akter},  and \cite{Alshomrani}, which combined U-Net variants with YOLOv2, CapsVGGNet, CNN, and VGG19 classifiers, respectively, with accuracies of 97\%, 99\%, and 98.7\% in the first three frameworks. However, the YOLOv2 classifier in \cite{KUMARSAHOO2023616} required 19.83 M parameters and 16 GB of GPU memory. In \cite{jabbar}, generalization and interpretability were not validated in the existing framework. Segmentation raised four-class accuracy only from 98.7\% to 98.8\% in \cite{akter}, and achieved a validation DSC of 75\% despite 98\% pixel accuracy in \cite{Alshomrani}, indicating overfitting. This redundant feature extraction added cost without a proportional classification gain. Multi-task designs removed redundant feature extraction and training at each stage by using a shared encoder, achieving DSCs of 0.92 and 0.86 with 97\% accuracy in \cite{KORDNOORI2024101931} and \cite{RABBY2024100346}, respectively. However, \cite{KORDNOORI2024101931} required both a mask and a label per image, whereas BT-Net \cite{RABBY2024100346} used separate datasets at the cost of a lower DSC. These frameworks captured only local features, missing global tumor information, and offered no interpretability. Accuracy-driven hybrids addressed these limitations at high computational cost, i.e., in \cite{renugadevi2025hybrid} fused CNN and Vision Transformer features with Grad-CAM, SHAP, and LIME XAIs, and the model in \cite{SALAMA2026433} combined CVAE-based synthesis, UNETR segmentation, and ResNet50--VGG16 classification, achieving DSC of 0.966 for WT and 97.45\% and accuracies of 99.18\% and 99.35\%, respectively. However, it requires $\approx$ 36.96 M and $\approx$ 109.4 M parameters, with the latter taking 1.65 s per image on an RTX 4090 GPU.

Deployment-oriented studies targeted distributed and low-resource settings. In \cite{ramprasad}, IoMT transmission was followed by RU-Net segmentation, BWO-GA feature selection, and AlexNet classification at the receiver, reporting 99.97\% segmentation and 99.98\% classification accuracy. However, its three independent modules increased complexity, and performance degraded when processing unseen images. The model was further processed centrally across all images. In \cite{neena2026haar}, a 2.8 M-parameter wavelet-attention CNN achieved 95.88\% four-class accuracy on a Raspberry Pi 5, but performed only classification and was trained centrally on a Tesla P100 GPU. Thus, the reviewed frameworks mainly performed joint segmentation and classification on GPU workstations, with limited attention to on-device execution and training on resource-constrained hardware, revealing the following gaps:
\begin{itemize}
    \item Redundant feature extraction in \cite{KUMARSAHOO2023616,jabbar,akter, Alshomrani} and heavy hybrid models in \cite{renugadevi2025hybrid,SALAMA2026433} exceed edge budgets. While the only edge model exempts the segmentation task in \cite{neena2026haar} and performs only the classification task. Hence, a single lightweight encoder serving both tasks is required.
    \item CNN-based multi-task models lost segmentation accuracy in \cite{KORDNOORI2024101931, RABBY2024100346}, while transformer hybrids regained it only at high cost in \cite{renugadevi2025hybrid,SALAMA2026433}. Hence, a decoder that recovers spatial and global context with minimal computation is required.
    \item Training was confined to GPU or cloud platforms for the models used in \cite{KUMARSAHOO2023616,Alshomrani,SALAMA2026433,neena2026haar}, and in \cite{ramprasad}, which sent raw images centrally. Hence, privacy-preserving training is required in which edge clients update part of the model locally and share only its weights.
    \item Evaluation was limited to GPU workstations in \cite{SALAMA2026433}, with resource usage reported only for a single task in \cite{neena2026haar}. Hence, on-device training and evaluation reporting CPU utilization, thermal stability, and latency is required.
\end{itemize}

\section{Overview of Proposed IoMT System for Brain Tumor Detection} \label{overview}
\begin{figure*}[!t]
    \centering
    \includegraphics[width=1\linewidth]{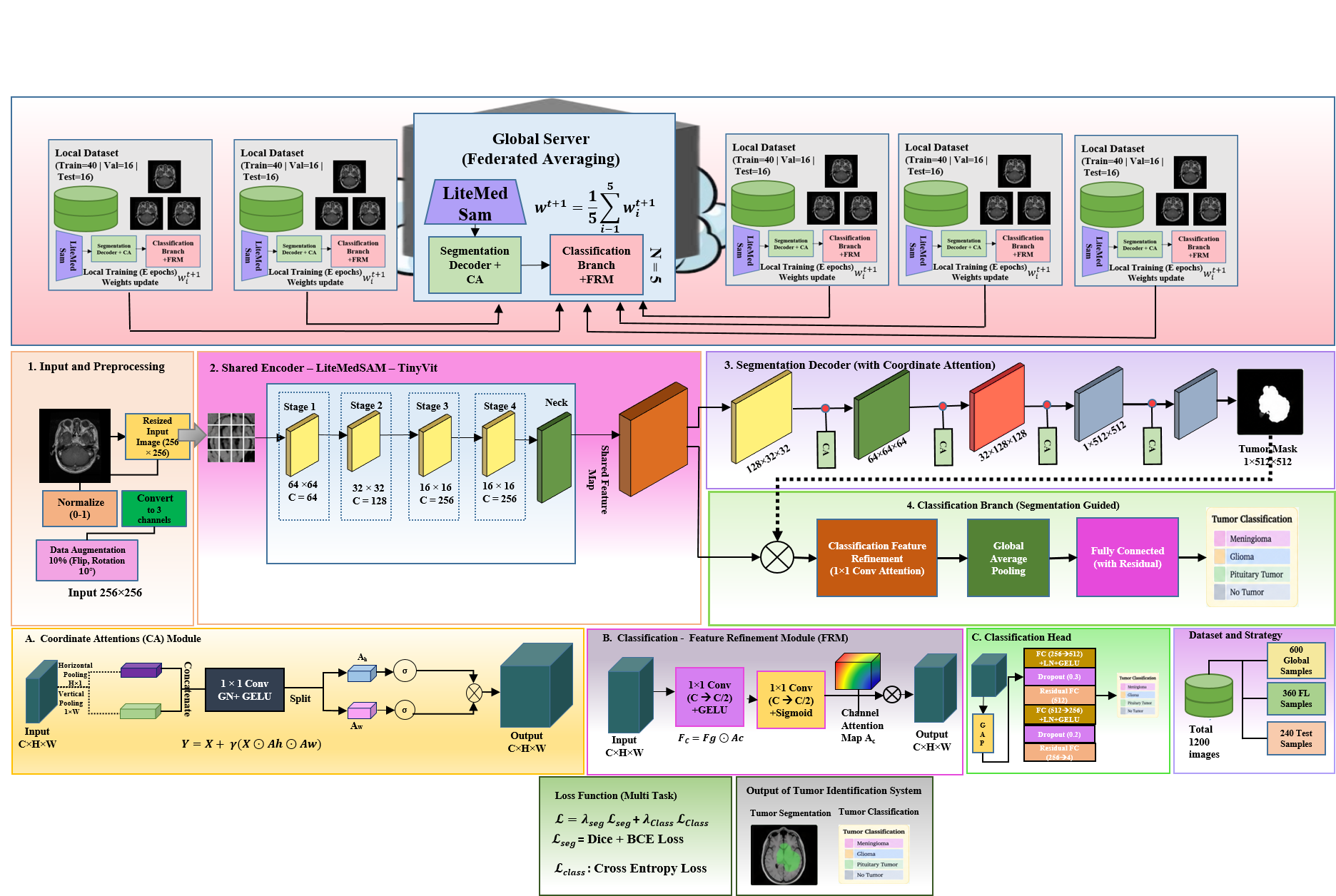}
    \caption{Overall Architecture of the proposed resource-constrained multi-task framework, illustrating the Federated Learning (FL) mechanism in the uppermost block and the input preprocessing pipeline in the second block of the first row.}
    \label{fig:proposed_framework}
\end{figure*}

The proposed framework is deployed as a two-tier FL-based IoMT system comprising a global server and $K$ edge clients, where each client represents a medical institution in a smart city holding a private MRI dataset $\mathcal{D}_i=\{(x_j,m_j,y_j)\}_{j=1}^{n_i}$, with $x_j$ the MRI image, $m_j$ the tumor mask, and $y_j$ the class label. In order to ensure patient privacy, raw images never leave the client, and only the model parameters $\boldsymbol{\theta}=\{\mathbf{W},\mathbf{b}\}$ are exchanged, where $\mathbf{W}$ and $\mathbf{b}$ represent weights and biases, respectively. Since segmentation and classification share a single encoder, each client optimizes one parameter set for both tasks by minimizing the joint loss expressed as:
\begin{equation}
	\mathcal{L}_i(\boldsymbol{\theta})=\frac{1}{n_i}\sum_{j=1}^{n_i}
	\Big[\lambda_s\,\mathcal{L}_{\mathrm{seg}}(S_j,m_j)+\lambda_c\,\mathcal{L}_{\mathrm{cls}}(\hat{y}_j,y_j)\Big],
	\label{eqn:fl_loss}
\end{equation}
where $\mathcal{L}_i(\boldsymbol{\theta})$ is the local loss of client $i$ for the model parameters $\boldsymbol{\theta}$. $n_i$ denotes the number of training samples at client $i$, and $j$ is the sample index. $\mathcal{L}_{\mathrm{seg}}$ and $\mathcal{L}_{\mathrm{cls}}$ denote the segmentation and cross-entropy classification losses, respectively. The weights $\lambda_s$ and $\lambda_c$ balance the segmentation and classification losses. In each communication round ($r$), client $i$ receives the global model $\boldsymbol{\theta}^{r-1}$ and performs local training on its dataset as follows:
\begin{equation}
	\boldsymbol{\theta}_i^{r}=\boldsymbol{\theta}^{r-1}-\eta\,\nabla\mathcal{L}_i(\boldsymbol{\theta}^{r-1}),
	\label{eqn:fl_local}
\end{equation}
where $\eta$ is the learning rate and $\nabla\mathcal{L}_i$ is the gradient of the local loss. The node executes $E$ local optimization epochs, updating only the designated trainable parameters while freezing the early encoder layers, thereby reducing on-device computational overhead. For experimental consistency, all clients are assumed to employ identical and fixed local training configurations, i.e., optimizer, learning rate $\eta$, batch size, and number of local epochs $E$ in every round, without client-specific hyperparameter adaptation. Thereby, isolating the effect of the proposed framework from variations in local optimization settings under heterogeneous deployment conditions \cite{tian2025federated}.The server further aggregates the client updates via Federated Averaging (FedAvg) using:
\begin{equation}
	\boldsymbol{\theta}^{r}=\sum_{i=1}^{K}\frac{n_i}{n}\,\boldsymbol{\theta}_i^{r},\qquad n=\sum_{i=1}^{K}n_i,
	\label{eqn:fl_agg}
\end{equation}
where $n_i$ denotes the local dataset size at client $i$ and $K$ is the number of participating clients. It reduces to a uniform average over $R$ communication rounds due to balanced client dataset sizes. The resulting cumulative communication overhead ($C_{\mathrm{comm}}$) is estimated using:
\begin{equation}
	C_{\mathrm{comm}}=R\cdot K\cdot\big(P_{\mathrm{down}}+P_{\mathrm{up}}\big),\qquad P\approx q\,|\boldsymbol{\theta}|,
	\label{eqn:fl_comm}
\end{equation}
where $P_{\mathrm{down}}$ and $P_{\mathrm{up}}$ denote the per-client downlink and uplink payload sizes, respectively. The global server–edge client communication link is assumed to be reliable, with no packet loss or retransmission delays; thus, the overhead in Eq. ~\eqref{eqn:fl_comm} depends solely on payload size. Each payload depends on the number of model parameters $|\boldsymbol{\theta}|$ and the number of bits $q$ used to represent each parameter. The client model jointly produces the segmentation output $S_j$ and class prediction $\hat{y}_j$ using a single parameter set $\boldsymbol{\theta}$, whose size determines the communication payload in Eq.~\eqref{eqn:fl_comm} and whose depth influences the local training cost in Eq.~\eqref{eqn:fl_local}. In order to meet the resource-constrained design criteria, the proposed IoMT framework integrates pixel-level tumor segmentation and four-class tumor classification into a single end-to-end architecture using a lightweight pretrained encoder. The proposed framework shares encoder features, reducing redundant computation and communication overhead by requiring each client to train and transmit only one parameter set, unlike two-stage approaches that train separate models for each task. The detailed architecture of the proposed framework is presented in Fig.~\ref{fig:proposed_framework}.

The proposed system is functionally decomposed into two major functional components: (i) a lightweight, partially frozen backbone encoder, and (ii) a shared-representation multi-task lightweight attention module decoder that performs segmentation and classification. The complete processing pipeline can be comprehended as:
\begin{equation}
I \rightarrow E(I) \rightarrow \{S, F_g\} \rightarrow \{S, C\},
\label{eqn:1}
\end{equation}
where $I$ denotes the input MRI image, $E(\cdot)$ represents the LiteMedSAM image encoder \cite{litemedsam}, $S$ represents the predicted tumor segmentation mask, $F_g$ denotes the segmentation-guided feature representation, and $C$ denotes the predicted tumor class. The algorithm used to perform the multi-task in a resource-constrained environment, as well as the training and validation procedures, are presented in Algorithm \ref{alg:proposed_framework}. 
\begin{algorithm}[t]
\caption{Resource-Constrained Federated Learning Trained Proposed Framework}
\label{alg:proposed_framework}
\small
\begin{algorithmic}[1]
\STATE \textbf{Input:} MRI dataset $\mathcal{D}$, pretrained LiteMedSAM
encoder $E_{\theta}$, number of clients $K=5$, communication rounds $R=5$
\STATE Initialize LiteMedSAM/TinyViT; freeze early encoder layers
and fine-tune the final two stages and neck
\STATE Build segmentation decoder $D_{\mathrm{seg}}$
($256\!\to\!128\!\to\!64\!\to\!32\!\to\!1$) with coordinate attention
\STATE Construct the classification head $H_{\mathrm{cls}}$
\STATE \textbf{Forward pass:}
$F=E_\theta(x)$, $S=D_{\mathrm{seg}}(F)$
\STATE $F_g=F\odot\operatorname{Resize}(\sigma(S))$
\STATE $A_c=\sigma(W_{c2}*\operatorname{GELU}(W_{c1}*F_g))$
\STATE $\hat{\mathbf y}=H_{\mathrm{cls}}
(\operatorname{GAP}(F_g\odot A_c))$
\STATE Train on global data to initialize and fine-tune $(W^0,b^0)$
\FOR{$r=1,\ldots,R$}
    \FOR{$i=1,\ldots,K$}
        \STATE Send $(W^{r-1},b^{r-1})$ to client $i$
        \STATE Locally train on $\mathcal{D}_i$ and return
        $(W_i^r,b_i^r)$
    \ENDFOR
    \STATE \textbf{FedAvg:}
    $(W^r,b^r)=\frac{1}{K}\sum_{i=1}^{K}(W_i^r,b_i^r)$
\ENDFOR
\STATE \textbf{Return:} Global model $(W^R,b^R)$
\end{algorithmic}
\end{algorithm}

\subsection{Lightweight Feature Extraction Encoder}
% The feature extraction stage is based on the LiteMedSAM, which uses a Lightweight TinyVit-based image encoder that offers a more computationally efficient architecture than large vision backbones \cite{litemedsam, tinyvit, aitken2021understandingencoderdecoderarchitecturesattend}. Rather than fine-tuning all encoder parameters, the proposed framework selectively freezes the layers of TinyVit and subsequently unfreezes the encoder neck. Therefore, the majority of the pretrained weights and biases are optimized during training. This selective fine-tuning strategy reduces the number of trainable parameters and limits unnecessary parameter updates. This strategy provides a more resource-efficient approach by reducing computational requirements without compromising the feature extraction for subsequent downstream tasks.

The feature extraction stage is based on LiteMedSAM, whose lightweight TinyViT-based image encoder is computationally more efficient than large vision backbones \cite{litemedsam, tinyvit, aitken2021understandingencoderdecoderarchitecturesattend} as represented in Fig. \ref{fig:proposed_framework}. Instead of fine-tuning all encoder parameters, the proposed framework selectively freezes TinyViT layers and unfreezes the encoder neck, so that the majority of the pretrained weights and biases are subsequently optimized during training. This selective fine-tuning reduces the number of trainable parameters and irrelevant weight updates, lowering computational requirements without compromising feature extraction for downstream tasks. The proposed lightweight encoder transforms the $256 \times 256$ input MRI into a compact high-level feature map.

Let the input image be
\begin{equation}
I \in \mathbb{R}^{3 \times 256 \times 256}.
\end{equation}

The image encoder produces an embedding
\begin{equation}
F = E_{\mathrm{TinyViT}}(I)
\label{eqn:2}
\end{equation}
where in Eq. \eqref{eqn:2} $F$ denotes the encoded feature map used simultaneously by the segmentation and classification pathways.

\subsection{Attentive Segmentation and Segmentation-Guided Feature-Enhanced Classification Decoder}
The decoder uses the shared feature maps extracted by the encoder to support both tumor segmentation and subsequent tumor classification.
\subsubsection{Lightweight Attention-Enhanced Segmentation}
% To enhance the segmentation task, lightweight coordination, the encoded feature maps are passed to a lightweight decoder for segmentation \cite{coordinateattention}. The decoder upsamples the feature maps and progressively reduces the number of channels at the progressive layer. The proposed framework utilizes lightweight coordinate attention to improve the spatial feature map. The pooled feature maps are combined and transformed through a 1$\times$1 lightweight pointwise convolution, 1D adaptive pooling, Group Normalization, and GELU activation, which makes it lightweight. The resulting feature map is separated into horizontal and vertical components and is projected back to the original channel dimension. Unlike conventional channel-only attention, the proposed attention block separately models spatial information along the height and width dimensions of the tumor region.

To enhance the segmentation, the encoded feature maps are passed to a lightweight decoder that upsamples them while progressively reducing the number of channels at each stage, as shown in Fig. \ref{fig:proposed_framework}. The lightweight coordinate attention \cite{coordinateattention} refines the spatial feature maps as represented in Fig. \ref{fig:proposed_framework}. The feature maps are pooled along the height and width dimensions using 1D adaptive pooling. It is then combined and transformed with a 1$\times$1 pointwise convolution with Group Normalization and GELU activation, keeping the block lightweight. The result is separated into horizontal as well as vertical components and projected back to the original channel dimension. Unlike conventional channel-only attention, this block separately models spatial information along the height and width of the tumor region. For an intermediate feature tensor
\begin{equation}
X \in \mathbb{R}^{B \times C \times H \times W},
\end{equation}
Two-directional representations are obtained using 1D adaptive average pooling is represented in Eq. \eqref{eqn:5} and Eq. \eqref{eqn:6} derived from \cite{coordinateattention}:
\begin{equation}
X_h = \operatorname{Pool}_h(X),
\label{eqn:5}
\end{equation}
\begin{equation}
X_w = \operatorname{Pool}_w(X).
\label{eqn:6}
\end{equation}

The transformed representation is subsequently separated into height and width components. The independent $1\times1$ convolutions followed by sigmoid activation are then used to generate the corresponding attention maps \cite{hu2018squeeze}, which are represented in Eq. \eqref{eqn:7} and Eq. \eqref{eqn:8}:
\begin{equation}
A_h = \sigma\left(\operatorname{Conv}_h(X_h)\right),
\label{eqn:7}
\end{equation}
\begin{equation}
A_w = \sigma\left(\operatorname{Conv}_w(X_w)\right).
\label{eqn:8}
\end{equation}

The directional attention maps \cite{zhang2019self} are combined multiplicatively as shown in Eq. \eqref{eqn:9} and Eq. \eqref{eqn:10}:
\begin{equation}
A = A_h \odot A_w,
\label{eqn:9}
\end{equation}
where $\odot$ denotes element-wise multiplication. A residual formulation is subsequently applied as
\begin{equation}
X_{\mathrm{out}} = X + \gamma\left(X \odot A\right),
\label{eqn:10}
\end{equation}
where $\gamma$ is a learnable scalar initialized to zero. This residual formulation enables the attention mechanism to progressively learn its contribution during training without disrupting the pretrained feature map at the start of optimization.

\subsubsection{Segmentation Guided Feature Extraction}
The predicted segmented tumor region is used to guide the classification pathway. First, the segmentation logits are converted into a probability map using the sigmoid function as shown in Eq. \eqref{eqn:11}:
\begin{equation}
M = \sigma(S).
\label{eqn:11}
\end{equation}

Since the segmentation decoder produces a higher-resolution feature map than the encoder, the predicted probability map $M$ is resized to $M_r$. The spatial dimensions of the encoder feature map using bilinear interpolation are represented in Eq. \ref{eqn:12}:
\begin{equation}
M_r = \operatorname{Resize}(M, H_F, W_F).
\label{eqn:12}
\end{equation}
where $H_F$ and $W_F$ are the height and width features of the predicted tumor region.

The resized segmentation probability map is then used as a spatial weighting mask for the encoder features as:
\begin{equation}
F_g = F \odot M_r,
\label{eqn:13}
\end{equation}
where in Eq. \eqref{eqn:13}  $F_g$ represents the segmentation-guided feature representation and $\odot$ denotes element-wise multiplication. Consequently, features corresponding to regions with high tumor probability receive greater emphasis, whereas features from regions with low predicted tumor probability are attenuated. This operation establishes an explicit information flow from the segmentation task to the classification task, enabling the classifier to make predictions using tumor-focused rather than globally distributed image features.

\subsubsection{Segmentation-Guided Feature Refinement for Classification}
The classification task reuses the encoded features along with the segmented tumor region for guided feature analysis. The segmented feature map spatially guides the features for class-based classification as shown in Fig. \ref{fig:proposed_framework}. From a resource-constraint perspective, the localized tumor region in the MRI image is also provided as input to the classification head. The segmentation-guided feature representation ($F_g$) is subsequently processed by a lightweight module consisting of two $1\times1$ convolutions with a GELU, which generates a channel-wise refinement map rather than expensive spatial convolutions. This refinement provides additional feature refinement capability without a substantial increase in computational complexity \cite{woo2018cbam, hendrycks2016gaussian}. The feature refinement module is represented mathematically in Eq. \eqref{eqn:14}:

\begin{equation}
A_c =
\sigma\left(
\operatorname{Conv}_{1\times1}^{(2)}
\left(
\operatorname{GELU}
\left(
\operatorname{Conv}_{1\times1}^{(1)}(F_g)
\right)
\right)
\right).
\label{eqn:14}
\end{equation}

The refined classification representation in Eq. \eqref{eqn:15} is subsequently utilized for final class prediction.

\begin{equation}
F_c = F_g \odot A_c,
\label{eqn:15}
\end{equation}
where $\odot$ denotes element-wise multiplication.

\subsubsection{Refined Feature Reuse for Classification Task}
The refined feature map $F_c$ is further converted to a compressed feature vector using adaptive average pooling as shown in Eq. \eqref{eqn:16} and as represented in Fig. \ref{fig:proposed_framework}. 

\begin{equation}
z = \mathrm{GAP}(F_c).
\label{eqn:16}
\end{equation}

The compact feature vector is passed through a Fully Connected(FC) classification head containing a linear transformation, layer normalization, GELU activation, dropout, and a residual FC block, as shown in Eq. \eqref{eqn:17}
\begin{equation}
z_r = z_1 + \mathcal{F}(z_1),
\label{eqn:17}
\end{equation}
where $\mathcal{F}(\cdot)$ represents the fully connected transformation followed by normalization, nonlinear activation, and dropout. The representation is subsequently projected to 256 dimensions and finally mapped to four class logits:
\begin{equation}
\hat{y} = W_2\phi(W_1z_r) + b,
\label{eqn:18}
\end{equation}
where in Eq. \eqref{eqn:18} $\phi(\cdot)$ represents the intermediate nonlinear transformation. The classification head used in the proposed framework substantially reduces the computational and memory overhead compared with classification heads that directly process high-resolution feature maps, as shown in Fig. \ref{fig:proposed_framework}.

\section{Results and Discussion} \label{result}
The proposed IoMT-based framework for brain tumor detection has been experimentally evaluated in order to investigate its classification performance, federated learning effectiveness, and computational feasibility under the considered experimental setup.
\subsection{Experimental Setup and Training Strategy}
\begin{figure} [!h]
    \centering
    \includegraphics[scale = 0.5]{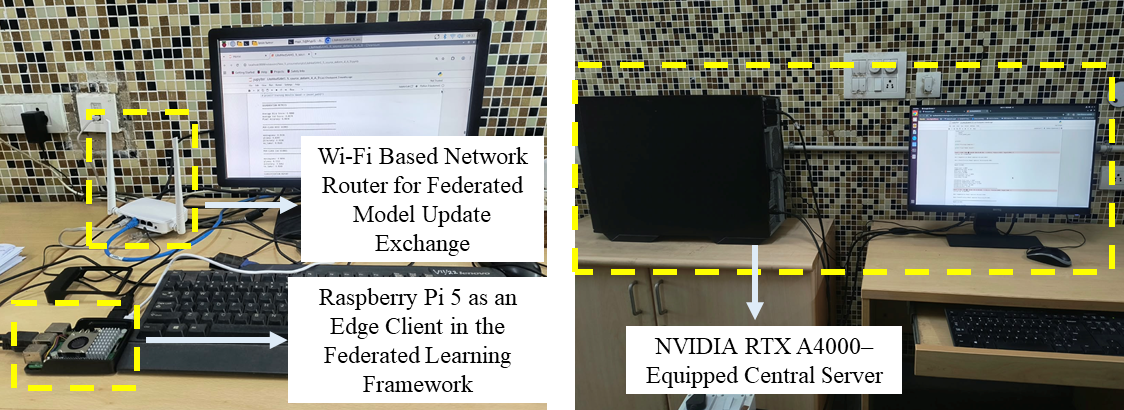}
    \caption{Experimental Setup of the Proposed IoMT-based Framework}
    \label{fig:expt-setup}
\end{figure}

The proposed IoMT-based framework has been implemented in a laboratory environment as shown in Fig. \ref{fig:expt-setup}. 
It is highlighted that the server-side computation has been executed on a workstation equipped with an Intel Xeon Silver 4214R CPU operating at 2.40 GHz with 12 cores, an NVIDIA RTX A4000 GPU with 16 GB memory, and 64 GB DDR4 RAM.

A total of 1200 class-balanced T1-weighted MRI samples are considered in this work, comprising Meningioma, Pituitary, and Glioma samples from the publicly available Figshare dataset \cite{figshare} and no-tumor samples from the Sartaj dataset \cite{sartaj}. Instead of a single centralized training configuration, the samples were divided into 600 for offline GPU-based global training, 360 for federated learning on resource-constrained clients, and 240 for testing, corresponding to 50\%, 30\%, and 20\%, respectively. The global training uses 480 samples for training, 60 for validation, and 60 for testing. The 360-sample client partition was sized to the compute and memory budget of the 16 GB Raspberry Pi 5 \cite{raspberrypi_documentation} and distributed equally across five clients, each holding 72 samples with a 40:16:16 samples for training:validation:test split. The remaining 240 samples form a held-out external test set shared by the global and federated configurations.

\subsection{Comparative Evaluation of Centralized and Distributed Learning Performance}
\begin{figure}[!h]
    \centering
\includegraphics[scale = 0.5]{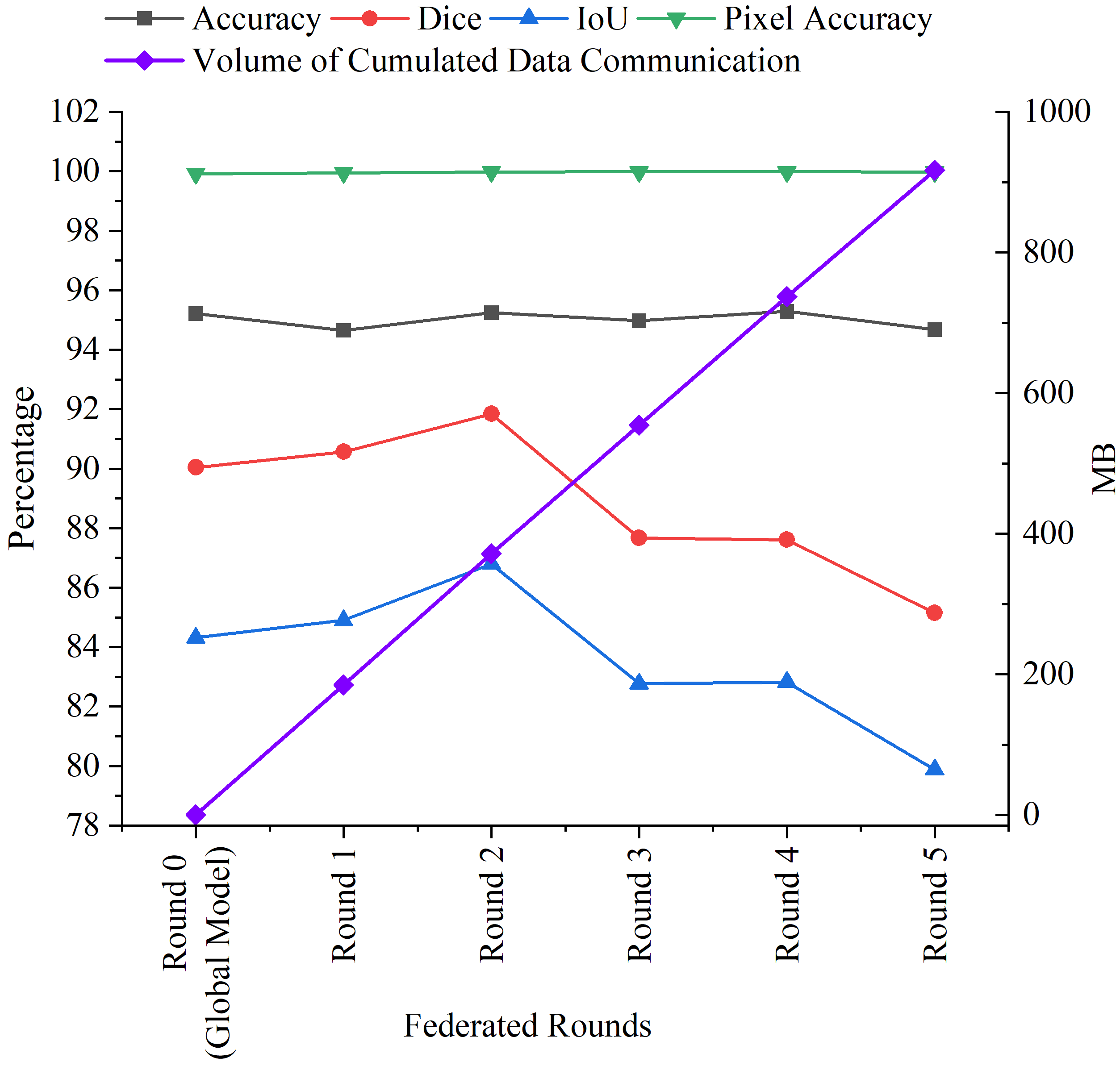}
    \caption{Performance Evaluation of the Proposed IoMT Framework Across Federated Rounds}
    \label{fig:placeholder}
\end{figure}

The client-based model is trained on GPU-accelerated weights and achieves classification accuracy close to the global baseline of 95.21\%, as shown in Fig.~\ref{fig:placeholder}. 
The accuracy remains close to the global baseline of 95.21\%, with relative changes ranging from a 0.59\% decrement to a 0.08\% increment across the five federated rounds.

% Relative to this baseline, accuracy decreases by 0.56, increases by 0.03, decreases by 0.24, increases by 0.08, and decreases by 0.54 in Rounds 1 to Round 5, respectively. 
It can be observed that the overall accuracy closely tracks the global baseline, varying from -0.59\% to +0.08\%. Round 2 achieves the highest segmentation fidelity, with relative improvements in DS and IoU of 2.00\% and 2.95\%, respectively. Subsequent rounds exhibit performance degradation across both overlap metrics despite pixel accuracy persisting above 99.9\%, highlighting its limited sensitivity to segmentation degradation. This indicates that local weight and bias updates on low-power hardware maintain performance close to that of the globally trained model. DSC and IoU exceed the global baselines of 90.04\% and 84.31\% in Round 1 and Round 2, peaking at 91.84\% and 86.80\%, which increases by 1.80\% and 2.49\%, respectively. Since both metrics move in the same direction in every round, this gain reflects a generalized improvement in weight updates. Round 2 provides the strongest performance, with accuracy of 95.24\%, DSC of 91.84\%, IoU of 86.80\%, and pixel accuracy of 99.98\%, all exceeding the global performance metrics by +0.03, +1.80, +2.49, and +0.06, respectively. 
%varying -0.05\% 
While pixel accuracy remains between 99.92\% and 99.99\% across all rounds, without highlighting substantial variation across rounds. DSC and IoU then decline to 85.15\% and 79.88\% by Round 5, showing that local weight updates adapt the shared feature representations to client-specific data distributions. Hence, the model potentially improves client-level adaptation and generalization rather than simply preserving the globally fine-tuned segmentation weights. Furthermore, the communication payload remains nearly constant at approximately 42.2 MB downlink and 48.7 MB uplink per round, varying by less than 0.05 MB, which amounts to 210.86 MB downlink and 243.65 MB uplink per client over five rounds, totaling 454.51MB. As a single multitask framework serves both segmentation and classification, separate payloads of comparable size for two independent pipelines are avoided, making the communication design suitable for resource-constrained edge clients.
% \begin{table}[htbp]
% \centering
% \caption{Segmentation Performance Across Federated Learning Rounds}
% \label{tab:federated_round_performance}
% \scriptsize
% \setlength{\tabcolsep}{5pt}
% \renewcommand{\arraystretch}{1.15}
% \begin{tabular}{lcccc}
% \hline
% \textbf{Round} & \textbf{Accuracy} & \textbf{Dice} & \textbf{IoU} & \textbf{Pixel Accuracy} \\
% \hline
% Global (50\% Samples) & 95.21\% & 90.04\% & 84.31\% & 99.92\% \\
% Round 1 & 94.65\% & 90.57\% & 84.90\% & 99.95\% \\
% Round 2 & 95.24\% & 91.84\% & 86.80\% & 99.98\% \\
% Round 3 & 94.97\% & 87.67\% & 82.77\% & 99.99\% \\
% Round 4 & 95.29\% & 87.61\% & 82.82\% & 99.99\% \\
% Round 5 & 94.67\% & 85.15\% & 79.88\% & 99.98\% \\
% \hline
% \textbf{240 Overall Testing} & \textbf{94.72\%} & \textbf{88.53\%} & \textbf{83.68\%} & \textbf{99.89\%} \\
% \hline
% \end{tabular}
% \end{table}

\subsection{Impact of Trainable Parameters on Model Performance and Hardware Efficiency}
In the proposed framework, the percentage of trainable parameters is systematically varied from 100\% to 70\%, as shown in Table \ref{tab:trainable_parameter_analysis}. Freezing 5\% of the parameters and utilizing 95\% trainable parameters decreases the number of trainable parameters by 489,565 and training FLOPs by 3.98 G, as shown in Table \ref{tab:trainable_parameter_analysis}. 
\begin{table}[!h]
\centering
\caption{Performance Evaluation Under Different Trainable Parameter Configurations}
\label{tab:trainable_parameter_analysis}
\scriptsize
\setlength{\tabcolsep}{4pt}
\renewcommand{\arraystretch}{1.1}

\begin{tabular}{c|c|c|c|c|c|c|c|c}
\hline
\textbf{Trainable (\%)} &
\textbf{Trainable Params.} &
\textbf{Frozen Params.} &
\textbf{FLOPs (Train/Total)} &
\textbf{Accuracy (\%)} &
\textbf{DSC (\%)} &
\textbf{IoU (\%)} &
\textbf{Pixel Acc. (\%)} &
\textbf{ROC-AUC (\%)} \\
\hline
100 & 9,791,300 & 0 & 79.55/79.55 G & 94.35 & 88.61 & 83.93 & 99.90 & 99.89 \\ \hline
95 & 9,301,735 & 489,565 & 75.57/79.55 G & 94.20 & 88.53 & 83.68 & 99.89 & 99.87 \\ \hline
90 & 8,812,170 & 979,130 & 71.60/79.55 G & 93.96 & 88.19 & 83.22 & 99.87 & 99.84 \\ \hline
85 & 8,322,605 & 1,468,695 & 67.62/79.55 G & 93.58 & 87.74 & 82.63 & 99.85 & 99.79 \\ \hline
70 & 6,853,910 & 2,937,390 & 55.69/79.55 G & 92.71 & 86.58 & 81.09 & 99.79 & 99.63 \\
\hline
\end{tabular}
\end{table}
%The highlighted text below is difficult to understand, specifically values ranged from, which value actually we are pointing to that we need to mention?
In comparison with the performance of 100\% trainable parameters, the 95\% trainable parameters exhibited changes in performance of -0.15\%, -0.08\%, -0.25\%, -0.01\%, and -0.02\% in Accuracy, DSC, IoU, Pixel Accuracy, and ROC-AUC, respectively. The 95\% to 90\% reduction in trainable parameters resulted in a variation of -0.24\% in accuracy, -0.34\% in DSC, and -0.46\% in IoU. At 70\% trainable parameters, the computational cost decreases to 55.69 G FLOPs, representing a reduction of 23.86 G FLOPs compared with 100\% trainable parameters. 
%The highlighted text below is difficult to understand, specifically cumulative performance part, is it accuracy/IoU or a specific metric we are talking about?
This configuration resulted in performance drop of 1.35\% in accuracy, 1.64\% in DSC, 2.33\% in IoU, 0.09\% in pixel-accuracy , and 0.22\% in ROC-AUC, compared with the 100\% trainable-parameter configuration. In comparison, the 95\% trainable configuration reduces the computational cost to 75.57 G FLOPs, with corresponding performance of -0.15\%, -0.08\%, -0.25\%, and -0.02\% in accuracy, DSC, IoU, and ROC-AUC, respectively.  This signifies that, with 95\% of parameters trainable, compared with 100\% trainable, it largely preserves distinct multiclass tumor features and performs multi-tasking effectively.
% ,about 490 K parameters and 4 GFLOPs while accuracy falls by only 0.15 pp Dice by 0.08 pp, and IoU by 0.25 pp. ROC-AUC remains unchanged by -0.02 pp, so the discriminative ability of the classifier is retained, and the small accuracy loss is a marginal threshold effect. The DSC change is also likely within acceptable variability, so the 95\% configuration is an effectively optimal choice for a lightweight multi-task parameter. The cost-benefit ratio gets worse as more layers are frozen. Accuracy loss per percentage of training FLOPs saved rises from about 0.030 pp at 95\% to 0.039pp at 90\%, 0.051pp at 85\%, and 0.055pp at 70\%, and IoU loss grows from 0.25 pp to 2.84 pp. This shows that the early layers perform crucial feature extraction, while freezing deeper layers in the network removes the features needed for segmentation and classification. 
The 95\% freezing configuration offers an optimal trade-off between computational efficiency and task-specific adaptation, while retaining most of the extracted features.

\subsection{Components-wise Ablation Analysis}
\begin{figure}[!h]
    \centering
    \includegraphics[scale= 0.45]{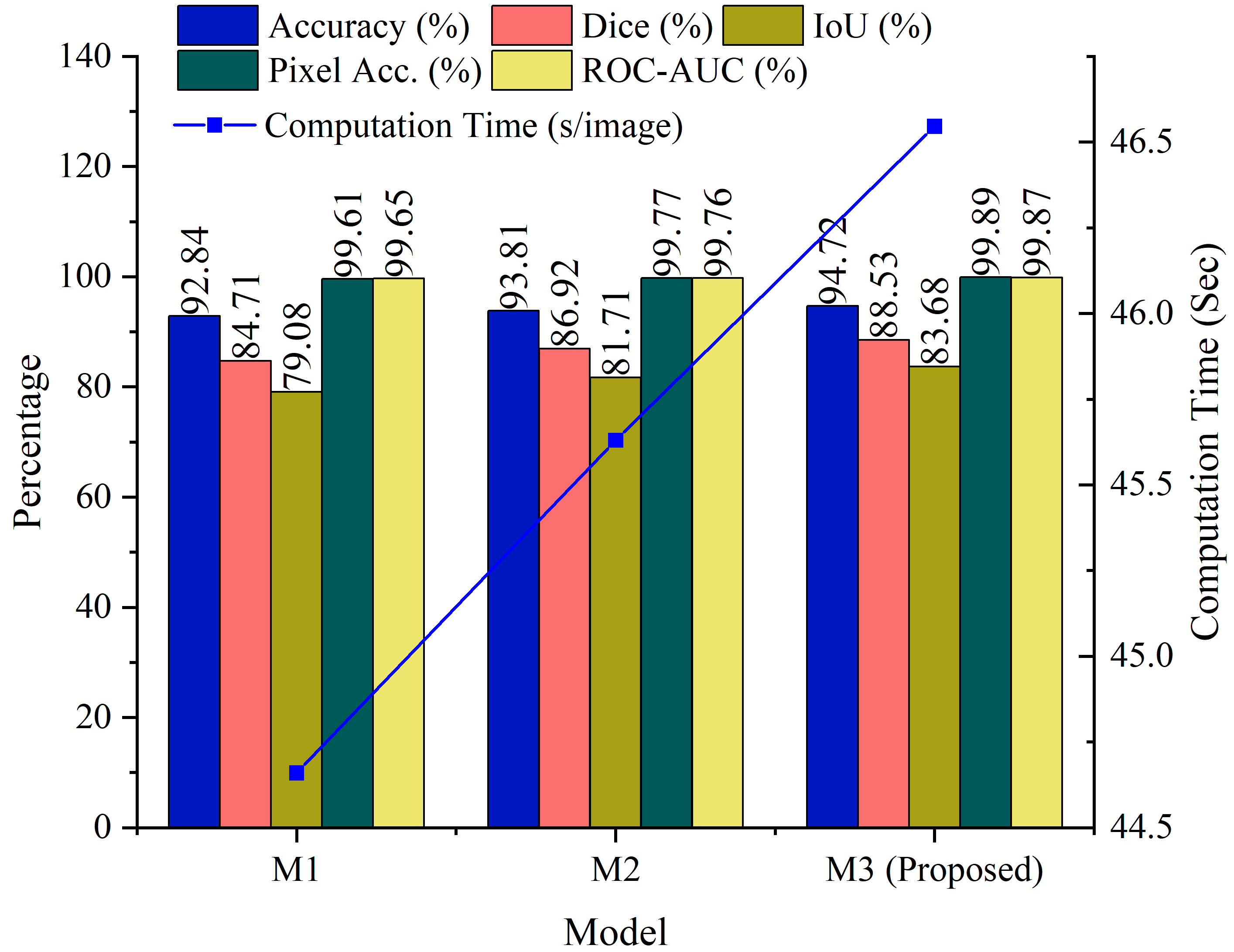}
    \caption{Ablation Study of the Proposed Framework}
    \label{fig:ablation}
\end{figure}

In order to determine the individual contribution of the proposed framework, the ablation study is conducted using three configurations: (i) M1: Fine-tuned LiteMedSAM + Classification Head; (ii) M2: M1+Coordinate Attention (CA) Module; (iii) M3: M2+Feature Refinement(FR) (Proposed Framework).
In Fig. \ref{fig:ablation}, the results demonstrate a progressive improvement as each component is added.
% The baseline M1 achieves an accuracy of 92.84\%, DSC of 84.71\%, and IoU of 79.08\%.In M2, adding the lightweight coordinate attention module to the baseline M1 improves accuracy by 0.97\%, DSC by 2.21\%, and IoU by 2.63\%. In M3, the addition of the feature refinement further improves accuracy by 0.91\%, DSC by 1.61\%, and IoU by 1.97\% compared to M2. In comparison with the baseline M1 and the proposed framework M3, overall improvements of 1.88\% in accuracy, 3.82\% in DSC, and 4.60\% in IoU are observed. Thus, the ablation results indicate that the additional architectural components contribute not only to multi-class classification but also improve the spatial segmentation of the tumor region. Although the computational complexity increases as the component progresses from M1 to M3, the proposed framework provides additional representational capability with a comparatively controlled computational overhead.
The baseline M1 achieves an accuracy of 92.84\%, DSC of 84.71\%, and IoU of 79.08\%. In M2, adding the lightweight coordinate attention module to the baseline M1 improves accuracy by 0.97\%, DSC by 2.21\%, and IoU by 2.63\%. In M3, the addition of the feature refinement further improves accuracy by 0.39\%, DSC by 1.61\%, and IoU by 1.97\% compared to M2. In comparison with the baseline M1 and the proposed framework M3, overall improvements of 1.36\% in accuracy, 3.82\% in DSC, and 4.60\% in IoU are observed. Thus, the ablation results indicate that the additional architectural components contribute not only to multi-class classification but also improve the spatial segmentation of the tumor region. Although the computational complexity increases as the component progresses from M1 to M3, the proposed framework provides additional representational capability with a comparatively controlled computational overhead.

These improvements are accompanied by a gradual increase in computational complexity and parameter count. The total FLOPs increase from 78.81~G in M1 to 79.18~G in M2 and 79.55~G in M3, while trainable parameters increase from 9,108,642 to 9,226,954 and 9,301,735, respectively. The corresponding CPU utilization rises from 50.7\% to 52.8\%, RAM usage from 6.88~GB to 7.01~GB, and CPU temperature from 58.9$^\circ$C to 60.6$^\circ$C. Subsequently, the proposed M3 introduces only 0.74~G in additional total FLOPs and 193,093 additional trainable parameters compared to M1, while improving DSC by 3.82\%, IoU by 4.60\%, and classification accuracy by 1.36\%. These results indicate that the additional computational cost is accompanied by a significant improvement in performance quality, with a comparatively modest increase in computational resources.

\subsection{Comparison with State-of-the-Art Frameworks on Edge Hardware}
Table~\ref{tab:combined_pi5_benchmark} compares the proposed IoMT framework with ten state-of-the-art brain tumor diagnosis frameworks, all evaluated on the same dataset using a 16~GB Raspberry Pi~5 client. 
\begin{table*}[!h]
\centering
\caption{Comparison of the Proposed Framework with State-of-the-Art Joint Segmentation and Classification Methods in Terms of Computational, Thermal, and Diagnostic Performance}
\label{tab:combined_pi5_benchmark}
\scriptsize
\setlength{\tabcolsep}{2.5pt}
\renewcommand{\arraystretch}{1.2}
\resizebox{\textwidth}{!}{%
\begin{tabular}{l|l|c|c|c|c|c|c|c|c|c|l|c|c|c|c|c}
\hline
\textbf{Author} &
\textbf{Model} &
\textbf{\makecell{CPU Util.\\(\%)}} &
\textbf{\makecell{RAM\\(GB)}} &
\textbf{\makecell{Memory BW\\Util. (\%)}} &
\textbf{\makecell{Peak Headroom\\(GB)}} &
\textbf{\makecell{Avg. Temp.\\($^\circ$C)}} &
\textbf{\makecell{Peak Temp.\\($^\circ$C)}} &
\textbf{Throttling} &
\textbf{\makecell{Latency\\(s/image)}} &
\textbf{\makecell{Round Time\\(min)}} &
\textbf{\makecell{BW Frequency /\\Execution Status}} &
\textbf{\makecell{Accuracy\\(\%)}} &
\textbf{\makecell{DSC\\(\%)}} &
\textbf{\makecell{IoU\\(\%)}} &
\textbf{\makecell{Pixel Accuracy\\(\%)}} &
\textbf{\makecell{ROC-AUC\\(\%)}} \\
\hline
Renugadevi (2025) \cite{renugadevi2025hybrid} &
\makecell[l]{VggVU-ED} &
92 & 13.2 & 82 & 2.62 & 78 & 85 & Yes &
380 & 354.67 &
\makecell[l]{2.2 GHz\\at Round 3} &
92.89 & 85.74 & 83.23 & 99.75 & 97.65 \\ \hline

NeenaKA (2026) \cite{neena2026haar} &
\makecell[l]{PWT + CNN +\\Self-Attention} &
68.5 & 4.25 & 35 & 11.57 & 58.6 & 64.2 & No &
350 & 326.67 &
\makecell[l]{2.09 GHz; Full run\\completed} &
93.42 & ** & ** & ** & 99.01 \\ \hline

Akter et al. (2024)\cite{akter} &
\makecell[l]{UNet +\\39-layer CNN} &
76 & 2.8 & 35 & 13.2 & 68 & 72 & Soft &
330 & 308 &
\makecell[l]{2.2 GHz; Clock reduced\\to 1.5 GHz at Round 3} &
93.58 & 83.45 & 80.98 & 97.75 & 97.89 \\ \hline

Kordnoori et al. (2024) \cite{KORDNOORI2024101931} &
\makecell[l]{Shared Encoder +\\U-Net Decoder + MLP\\(Global)} &
87 & 12.28 & 80 & 3.54 & 82 & 85 & Yes &
-- & -- &
\makecell[l]{2.4 GHz; Stopped\\during Round 1} &
97.69 & 95.82 & 93.89 & 99.85 & 99.79 \\ \hline

Salama (2026) \cite{SALAMA2026433} &
\makecell[l]{CVAE-UNETR-ResNet50\\+ VGG16 (Global)} &
98 & 15.6 & 98 & 0.2 & 83 & 85 & Yes &
-- & -- &
\makecell[l]{2.4 GHz; Stopped\\during Round 1} &
98.56 & 96.82 & 94.82 & 99.92 & 99.85 \\ \hline

Ramprasad et al. (2023) \cite{ramprasad} &
\makecell[l]{SBTC-Net + Black Widow\\+ Genetic Optimization} &
85 & 9.5 & 89 & 6.32 & 73 & 82 & Yes &
300 & 280 &
\makecell[l]{2.2 GHz; Clock reduced\\to 1.5 GHz at Round 2} &
93.21 & 85.62 & 82.64 & 97.89 & 98.98 \\ \hline

Jabbar et al. \cite{jabbar} &
\makecell[l]{CapsNet + VGGNet} &
94.2 & 7.80 & 68.5 & 8.02 & 74.2 & 82.1 & Yes &
456 & 426 &
\makecell[l]{2.2 GHz; Clock reduced\\to 1.5 GHz at Round 4} &
92.98 & 86.59 & 83.02 & 98.35 & 98.58 \\ \hline

Alshomrani et al. \cite{Alshomrani}&
\makecell[l]{Residual U-Net +\\FT-CNN (Global)} &
94.5 & 10.2 & 98 & 5.62 & 83 & 85 & Yes &
-- & -- &
\makecell[l]{2.4 GHz; Execution\\stopped} &
95.29 & 94.77 & 93.54 & 97.99 & 98.99 \\ \hline

Sahoo et al. (2023) \cite{KUMARSAHOO2023616} &
\makecell[l]{ResNet-18 U-Net\\+ YOLOv2 (Global)} &
99 & 7.8 & 68 & 8.0 & 83 & 85 & Yes &
-- & -- &
\makecell[l]{2.4 GHz; Stopped\\during Round 1} &
96.72 & 88.92 & 85.79 & 99.78 & 100.00 \\ \hline

Rabby et al. (2024) \cite{RABBY2024100346} &
\makecell[l]{BT-Net} &
91 & 8.9 & 64 & 6.9 & 75 & 79 & No &
490 & 457.34 &
\makecell[l]{2.0 GHz; Full run\\completed} &
91.89 & 86.92 & 81.99 & 95.67 & 97.71 \\ \hline

\hline
\textbf{Proposed Framework} &
\textbf{FedTumor-IoMT} &
\textbf{52.8} & \textbf{7.01} & \textbf{55} &
\textbf{8.81} & \textbf{60} & \textbf{65} &
\textbf{No} & \textbf{46.54} & \textbf{45} &
\makecell[l]{\textbf{1.5 GHz; Full run}\\
\textbf{completed}} &
\textbf{94.20} & \textbf{88.53} & \textbf{83.68} &
\textbf{99.89} & \textbf{99.87} \\

\hline
\end{tabular}%
}
\vspace{2mm}
\begin{minipage}{\textwidth}
\raggedright
\footnotesize
\textit{Note:} ``--'' in the Latency and Round Time columns denotes frameworks marked (Global), which are trained and evaluated on GPU-based global systems. The per-image latency and round time are therefore not reported, as they are not directly comparable with those of the resource-constrained edge clients. The `**' in the DSC, IoU, and Pixel Accuracy columns indicates that the framework performs only classification and does not report segmentation metrics.
\end{minipage}
\end{table*}
Where available, the reported hyperparameters of each framework are adopted; otherwise, a common baseline of batch size $=2$, learning rate $=0.001$, and the Adam optimizer is used to maintain experimental consistency. The comparison covers diagnostic performance, thermal behavior, and computational efficiency.

\subsubsection{Diagnostic Performance}
Existing frameworks largely rely on separately trained segmentation and classification networks, in which the classifier does not explicitly exploit the predicted tumor region. Under edge execution, the existing frameworks achieve accuracies of only 91.89--93.58\%, and those performing segmentation achieve DSC values of 83.45--86.92\% \cite{akter, renugadevi2025hybrid, neena2026haar, jabbar, RABBY2024100346, ramprasad}. In contrast, the proposed framework addresses this repetitive training through segmentation-guided feature reuse, achieving 94.20\% accuracy, 88.53\% DSC, 83.68\% IoU, 99.89\% pixel accuracy, and 99.87\% ROC-AUC, the highest values on all five metrics among the frameworks executed on the edge client. The shared encoder with U-Net decoder and MLP classifier \cite{KORDNOORI2024101931}, CVAE-UNETR-ResNet50-VGG16 \cite{SALAMA2026433}, Residual U-Net+FT-CNN \cite{Alshomrani}, and ResNet-18 U-Net+YOLOv2 \cite{KUMARSAHOO2023616} report higher accuracies of 95.29--98.56\%. However, these values reflect a global GPU-based evaluation, since the resource demands of these architectures could not be accommodated by resource-constrained training on edge devices. Hence, the proposed framework provides competitive, effective diagnostic performance in comparison with GPU-scale models while completing all federated rounds on the edge client.

\subsubsection{Thermal Behavior}
ResNet-18 U-Net+YOLOv2 \cite{KUMARSAHOO2023616}, VggVU-ED \cite{renugadevi2025hybrid}, CVAE-UNETR-ResNet50-VGG16 \cite{SALAMA2026433}, Residual U-Net+FT-CNN \cite{Alshomrani}, and the shared encoder model in \cite{KORDNOORI2024101931} reach peak temperatures of 85$^\circ$C, while CapsNet+VGG \cite{jabbar} and SBTC-Net \cite{ramprasad} reach 82.1$^\circ$C and 82$^\circ$C, respectively. In UNet+39-layer CNN \cite{akter}, CapsNet+VGG \cite{jabbar}, and SBTC-Net \cite{ramprasad}, the processor frequency is reduced to 1.5~GHz in later rounds, lowering the effective computational capacity and extending training time. The proposed framework extracts features once and reuses them for both tasks, utilizing a partially frozen shared encoder. This, in contrast, limits the sustained processor load, maintaining an average temperature of 60$^\circ$C and a peak of 65$^\circ$C at a fixed 1.5~GHz clock, with no throttling events across all five federated rounds. The proposed framework achieves thermal performance comparable to that of PWT+CNN+Self-Attention \cite{neena2026haar}. This existing architecture performs only the classification task and skips the segmentation task, with average and peak temperatures of 58.6$^\circ$C and 64.2$^\circ$C. In contrast, the proposed framework jointly performs segmentation and classification tasks, with each round 7.3 times faster at slightly higher average and peak temperatures of 60 $^\circ$C and 65$^\circ$C, respectively.

\subsubsection{Computational Efficiency}
VggVU-ED \cite{renugadevi2025hybrid}, CapsNet+VGG \cite{jabbar}, Residual U-Net+FT-CNN \cite{Alshomrani}, CVAE-UNETR-ResNet50-VGG16 \cite{SALAMA2026433}, and ResNet-18 U-Net+YOLOv2 \cite{KUMARSAHOO2023616} operate at 92--99\% CPU utilization, leaving insufficient headroom for gradient computation and federated communication due to the redundant feature extraction across each separate network, saturating the processor computational memory.
% Memory demand imposes a further constraint. 
VggVU-ED \cite{renugadevi2025hybrid} requires 13.2~GB of RAM, and its training process is terminated by the operating system at Round~3, due to memory exhaustion. CVAE-UNETR-ResNet50-VGG16 \cite{SALAMA2026433} requires 15.6~GB of RAM with only 0.2~GB of headroom, while CVAE-UNETR-ResNet50-VGG16 \cite{SALAMA2026433} and Residual U-Net+FT-CNN \cite{Alshomrani} consume 98\% of the memory bandwidth, and SBTC-Net \cite{ramprasad} consumes 89\%. As these architectures are designed for GPU workstations, their memory and processing requirements exceed the budget of an edge-device-based, resource-constrained client. The models in \cite{KORDNOORI2024101931, SALAMA2026433, Alshomrani, KUMARSAHOO2023616} exceed the on-device training budget beyond the first round. In contrast, the proposed framework addresses this by using a joint segmentation and classification approach with a single shared encoder. This eliminates the need for redundant training for feature extraction, reducing the number of trainable parameters. It operates at 52.8\% CPU utilization, 7.01~GB of RAM with 8.81~GB of headroom, and 55\% memory bandwidth utilization, leaving sufficient resources for local training and weight exchange among the global server and client-based server.

A low memory footprint alone, however, does not ensure efficient execution. PWT+CNN+Self-Attention \cite{neena2026haar} uses only 4.25~GB of RAM and 35\% memory bandwidth, yet requires 350~s/image and 326.67~min/round, while BT-Net \cite{RABBY2024100346} requires 490~s/image and 457.34~min/round. In contrast, the proposed framework performs the entire brain tumor diagnosis by sharing encoder feature maps between segmentation and classification and processing them through a lightweight decoder and classification head, reducing the per-image processing time to 46.54~s and the round time to 45~min. Consequently, the proposed framework achieves a per-image processing time 6.4 times lower than that of SBTC-Net \cite{ramprasad} and 10.5 times lower than that of BT-Net \cite{RABBY2024100346}, which represent the fastest and slowest existing frameworks with measurable latency, respectively. Likewise, the joint task reduces the round time by 7.3 and 10.2 times, respectively, as compared to PWT+CNN+Self-Attention \cite{neena2026haar} and BT-Net \cite{RABBY2024100346}, which complete all federated rounds without throttling.

Overall, the existing frameworks are constrained on the edge client by limited processing and memory resources, elevated temperatures, or prolonged round times. The proposed framework addresses each of these through shared feature extraction, segmentation-guided feature reuse and refinement, and selective encoder freezing. Consequently, it provides stable, multi-round, client-based federated training with the highest diagnostic performance among edge-executed frameworks.

\section{Conclusion}  \label{conclusion}
This work experimentally validates the feasibility of joint brain tumor segmentation and classification on resource-constrained edge devices within an FL-based IoMT system. The proposed architecture eliminates redundant feature extraction by sharing a single LiteMedSAM encoder and reusing segmentation-guided features. It completes all five federated rounds without thermal throttling, achieving 94.20\% classification accuracy and 88.53\% DSC, the highest among frameworks executed on the same hardware, while its accuracy across the five federated rounds remains within 0.6\% of the GPU-trained baseline. The coordinate attention and feature refinement modules improve DSC and IoU by 3.82\% and 4.60\%, respectively, with only 0.74 GFLOPs of additional computation. Compared with existing frameworks, the proposed approach achieves 52.8\% CPU utilization and 6.4--10.5 times lower per-image processing time. Training 95\% of the parameters reduces computation by 3.98 GFLOPs with at most 0.25\% degradation in any metric.

In future work, the proposed framework will be extended from T1-weighted MRI to multimodal MRI with tumor sub-region segmentation, and from five balanced clients to larger client populations with non-identical data distributions across diverse edge platforms. In addition, model quantization and secure aggregation will be investigated to further reduce computational overhead and strengthen privacy during federated weight exchange.

\bibliographystyle{IEEEtran}
\bibliography{jbhi}  %%% Uncomment this line and comment out the ``thebibliography'' section below to use the external .bib file (using bibtex) .





% \end{thebibliography}

\end{document}